\documentclass[conference]{IEEEtran}
\usepackage{graphicx}
\usepackage{xspace}
\usepackage{amssymb}
\usepackage{bm}
\usepackage{amsmath}
\usepackage[sorting=none]{biblatex} %Imports biblatex package
\ifCLASSINFOpdf
\else
\fi

\begin{document}
\title{Data optimization for large batch distributed training of deep neural networks}

\author{\IEEEauthorblockN{Shubhankar Gahlot}
\IEEEauthorblockA{Oak Ridge National Lab\\
Oak Ridge, USA\\
Email: sgahlot@hawk.iit.edu}
\and
\IEEEauthorblockN{Junqi Yin}
\IEEEauthorblockA{Oak Ridge National Lab\\
Oak Ridge, USA\\
Email: yinj@ornl.gov}
\and
\IEEEauthorblockN{Mallikarjun (Arjun) Shankar}
\IEEEauthorblockA{Oak Ridge National Lab\\
Oak Ridge, USA\\
Email: shankarm@ornl.gov}}


%Telephone: (800) 555--1212, Fax: (888) 555--1212}


\maketitle

\begin{abstract}
% OK to have a broad context setting line here.. 
Distributed training in deep learning (DL) is common practice as data and models grow. 
The current practice for distributed training of deep neural networks faces the challenges of communication bottlenecks when operating at scale, and model accuracy deterioration with an increase in global batch size.  Present solutions focus on improving  message exchange efficiency as well as implementing techniques to tweak batch sizes and models in the training process. The loss of training accuracy typically happens because the loss function gets trapped in a local minima. We observe that the loss landscape minimization is shaped by both the model and training data and propose a data optimization approach that utilizes machine learning to implicitly smooth out the loss landscape resulting in fewer local minima. Our approach filters out data points which are less important to feature learning, enabling us to speed up the training of models on larger batch sizes to improved accuracy.
\end{abstract}

% Add keywords

\section{Introduction} \label{sec:introduction}
Distributed strategies for scaling DL have become the cornerstone of solutions to the problems that arise due to ever growing data and model sizes. Scaling a DL workload requires careful calibration of compute, I/O, and communication among the compute nodes \cite{8945109}. In addition, we need to use large batch sizes to achieve weak scaling to make full use of the parallel hardware. We achieve training speedup by increasing the overall throughput of the system with fewer updates of individual model replicas. Distributed strategies, however, present a set of challenges \cite{1812.06162} that arise when using large batches across distributed computing platforms:

\textbf{Challenge 1:} Larger mini-batch sizes average the gradients for large samples thus reducing the variance and providing better estimate of model weights \cite{1711.00489}. The goal is to take bigger step sizes and in turn make the optimization algorithm progress faster. However, as shown in \cite{1705.08741}, increasing the mini-batch size leads to a drop in accuracy and leads to the generalization gap - difference in accuracy between training and testing.

\textbf{Challenge 2:} When using large clusters, it is harder to achieve near-linear scalability as the number of compute nodes increases, especially for models with a high communication bandwidth requirement. In both data parallel and model parallel distributed strategies, the communication step usually becomes the bottleneck as the number of GPUs increases. To improve performance for such distributed training systems, we need faster compute, more efficient bandwidth utilization, and more efficient collective primitives that can handle a system with thousands of GPUs.

It has been shown \cite{1812.06162, visualizeloss} that the loss landscape is a function of  both the DL model and the training data. Current scaling techniques focus more on parallelizing the workloads through data parallel training, but they suffer at large scale as the training batch size grows. This is because the likelihood of the optimization process being stuck in a local minima increases with increasing batch size. Our goal is to empirically identify techniques to effectively smooth the loss landscape and also develop guidelines for optimizing batch sizes for training. We address this by exploring the effect of batch size on model convergence and accuracy. Our experiments apply to ResNet models using the CIFAR-10 ~\cite{cifar10} data set. Our contributions are as follows:

\begin{itemize}
\item We present empirical evidence for the non-linear correlation between batch size and model accuracy for increasing batch sizes from 16 to 20,480. We provide recommendations for the appropriate batch sizes to use for the particular ResNet models.
\item We develop techniques using machine learning (ML) to empirically characterize data quality, and propose a data optimization approach to model scaling. We  cluster  data  points  using  ML and remove  those  classified  as noise. We use the remaining data to train another network (while keeping  the  testing  data  unchanged)  to equivalent and even higher levels of accuracy for certain large batch sizes thus saving system bandwidth and training time.

\item We compare the scaling efficiency for ResNet models between PyTorch's built-in distributed data parallel (DDP) and the Horovod \cite{horovod} communication library for up to 1536 V100 GPUs and find that the overheads of the dynamic messaging queue in Horovod get in the way of achieving the benefits expected from improved overlapping of communication and computation.   
\end{itemize}

The rest of the paper is structured as follows. We discuss the background and context for our work in Section~\ref{sec:background}. We present the details of our methodology and experiments in Section ~\ref{sec:methodology}. Section ~\ref{sec:results} discusses our results along with findings and recommendations. We conclude in Section ~\ref{sec:conclusion} with a discussion of future extensions.

\section{Background and Context for Data-optimized Scaling} \label{sec:background}
\subsection{Background} We list a few concepts necessary for the description that follows.

\subsubsection*{Dimensionality reduction} Dimensionality reduction is the process of transforming data from a high-dimensional space into a low-dimensional space so that the low-dimensional representation retains some meaningful properties of the original data, ideally close to its intrinsic dimension.

\subsubsection*{Feature extraction using CNN} Feature extraction may be deemed a data reduction technique. In a CNN the output of a convolutional layer captures features (feature maps) of the input data. The last convolutional layer captures the finest features (edges, curves, etc) from the data that it is trained on. Hence, we are interested in the output of that last convolutional layer and this process in general is called feature extraction using a CNN \cite{9012507}.

\subsubsection*{t-distributed stochastic neighbor embedding (t-SNE)}
t-SNE is a technique for dimensionality reduction that is particularly well suited for the visualization of high-dimensional data sets \cite{tsne}. 
\subsubsection*{Clustering}

Cluster analysis or clustering is an unsupervised machine learning technique in which more similar (based on some metric) data points group together to form a cluster. There are various criteria for clustering. %Some common clustering algorithms are listed in Table~\ref{tb:cluster}.
We use DBSCAN \cite{dbscan} for clustering data points in feature space because it is a density based technique and therefore doesn't assume any apriori distribution and can be scaled to large number of data points easily.

%\begin{table}
%\vspace{0.1cm}
%\caption{Common clustering algorithms}\label{tb:cluster}
%\begin{tabular}{ |p{3.75cm}||p{3.75cm}|  }
% \hline
% Clustering algorithm & Metric \\
% \hline\hline
% K-Means 
% 
% (centriod-based) & Distances between points\\
% \hline
% DBSCAN 
% 
% (density-based) &   Distances between nearest points  \\
% \hline
% Agglomerative 
% 
% (hierarchy-based) & Any pairwise distance\\
% \hline
%\end{tabular}
%\vspace{0.1cm}
%\end{table}
\subsection{Data Quality Context and Scaling Objectives}

\begin{itemize}
    \item{\it{Defining data quality:}} While there has been some work on relating physical attributes of data sets such as data set equilibrium, size, quality of label and contamination to model quality \cite{1906.11882}, there has been little to no significant work on how intrinsic properties of a data set affect model training performance. A central reason is because data quality is defined differently for different types of data - ranging widely from domain specific to broad information theoretic measures. For example, quality metrics for traditional image data are different for speech signal data or other types of higher order spectral image data. Our approach is to  use machine learning to filter out data points which are not relevant to training effectiveness. This gives us a way of thinking about how the quality of data elements affect the training outcomes. We use unsupervised techniques like dimensionality reduction and clustering to categorize data points and filter out those we classify as "noise" in feature space. Various dimensionality reduction, clustering, and data sampling techniques may be used for this step with different degrees of effectiveness.
    
    \item{\it{Feature (space) noise vs. Gradient descent noise:}} Feature (space) noise refers to the kinks and troughs in the loss landscape as described in \cite{visualizeloss} that can hinder the optimizer to advance towards the global minima whereas gradient descent noise refers to the noise in stochastic gradient descent (SGD) which gives the gradient descent its stochastic nature. It is known that as the batch size increases, stochasticity decreases, and therefore we can improve generalization either by adding noise to the SGD or by smoothing out the loss landscape. (This is similar to the role of skip connections in ResNets \cite{visualizeloss}.) 
    
    \item{\it{Smoothing the loss landscape:}} The loss landscape is shaped by the model as well as by the data. The higher dimension "topography" of the loss function determines how well the model will converge and to what accuracy. There can exist kinks and twirls in the loss landscape due to data points irrelevant to feature learning, which can make the model convergence a challenge especially at large batch sizes. If we can identify and remove those data points, the loss landscape would become smoother and easier for the model to move towards convergence. By classifying noise in feature space and removing them from the data set, we aim to speed up convergence.
    \item{\it{Distributed large batch training:}} There has been significant work applying data distribution to scale deep learning workloads. Models using batch sizes of 64k have been trained in under 4 mins \cite{1807.11205} using techniques such as Layer-wise Adaptive Rate Scaling (LARS) \cite{1708.03888} and Adaptive Batch Sizing. However, we focus our attention on scaling from an empirical, intrinsic data quality (or effectiveness) perspective. 
\end{itemize}

In the work we describe here, we take a data optimization approach to model scaling and connect the effectiveness of a data item to the features it enables in a trained model. We remove the data points classified as noise to feature learning and retrain a new model at scale. Our hypothesis is that clustering (and/or sampling) and removing data points classified as noise in feature space will filter data in such a way that it will smooth out the loss landscape. This will facilitate gradient traversal and thereby accelerate model convergence. 
%However, instead of using DBSCAN for classifying noise and reducing data other clustering techniques can be explored. Also %instead of using a clustering technique, data can be be sampled by spatial stratified sampling as well which has been left %as a future exercise.

\begin{figure*}[ht]
    \centering
    \includegraphics[width=18cm]{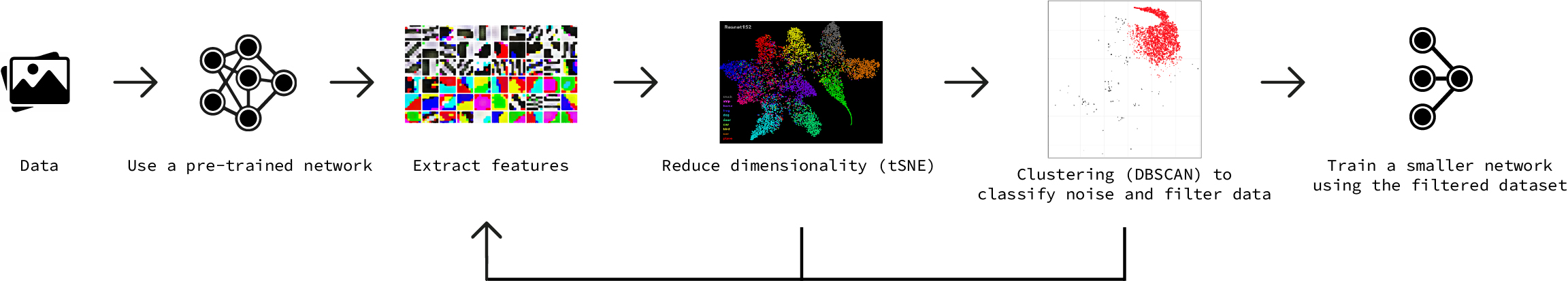}
    \centering
    \caption{Steps showing the process of applying data optimized strategy by (1) fine tuning a pretrained network, (2) Extracting and reducing dimensionality by PCA / t-SNE, (3) Clustering using DBSCAN to remove noise and (4) Training another network faster and at improved accuracy. }
    \label{fig:process}
\end{figure*}

\section{Methodology and Experiments} 
\label{sec:methodology}
\subsection{Conceptual Overview} 
We use a pretrained network and treat data quality as the output of a machine learning technique which can be used to remove data elements less significant to training. This helps smooth the loss landscape and thus acts as a data-optimized approach to distributed training and model scaling. Figure ~\ref{fig:process} outlines this process. The pipeline begins at the left by processing the data in a pretrained network. Next, the data elements that apply to the primary clusters of the penultimate ($n-1$th) layer are compactly represented through dimensionality reduction (with PCA  and t-SNE). These data elements are  clustered using DBSCAN to remove what we consider ``noise''. Finally, we train another smaller network, faster, and at improved accuracy.

More formally, let us represent CIFAR-10 data set by a matrix as follows:

$${\bf{X_{50k}} \in \mathbb{R}^{{50000 {\times} 3 {\times} 28 {\times} 28}}} \space - \space(1)$$

We represent a batch as follows: $${\bf{X_{b}} \in \mathbb{R}^{{b \times 3 \times 28 \times 28}}} \space - \space(2)$$ ({\it{NCHW}}) where {\it{N} or \bf{b}} is the batch size. Now let's define the feature extractor neural network (NN) by

$${\textbf{y}}=f({\textbf{x},\bm{\theta}}) \space - \space(3)$$

where $\bf{x}\in\mathbb{R}^{{3\times28\times28}}$ is one of the image examples and $\bm{\theta}$ is the value of parameters NN learns that result in the best function approximation. Since $\bf{y}$ is the set of features and output of the penultimate layer therefore,

$${\bf{y}\in
\begin{cases}
{\mathbb{R}^{512 \times 1} \quad \textrm{for ResNet-18 and 34,} \quad}\\
{\mathbb{R}^{2048\times1} \quad \textrm{for ResNet-50, 101 and 152} \quad }
\end{cases}  }\space-\space(4)$$
Now, we further reduce the dimensionality of $\bf{y}$ using t-SNE $t$.

$${\bf{y^{'}}}=t({\bf{y}},\bm\phi)-\space(5)$$
where $t$ maps $\bf{y}\to\bf{y^{'}}$ $\in\mathbb{R}^{2\times1}$ and is parameterized on $\bm\phi$.
Next, we perform clustering on each class individually using DBSCAN algorithm to classify and filter out noise. Let $d$ denote the DBSCAN function, parameterized on $\bm\gamma$, that maps $\bf{y^{'}_c}\to\bf{y^{'}_{c\alpha}}$ where $c\in[1,2,3,4,5,6,7,8,9,10]$ denotes individual class label and ${{\alpha}}\in[0,1]$, $0$ for noise and $1$ otherwise.

$${{\bf{y^{'}_{c\alpha}}}=d(\bf{y^{'}_c},\bm\gamma) \space}-\space(6)$$

Therefore, $\bf{y^{'}_{c0}}$  and $\bf{y^{'}_{c1}}$ denote data points classified as noise and not noise respectively for class $\bf{c}$. Finally, we combine all images i.e. ${\bf{y^{'}_{c1}}}s$, resulted from filtering and train a smaller network.

\subsection{Process}
We expand on the data set details for the steps listed above.
\subsubsection{Initial training:} First, we train ResNet-152, 101, 50, 34 and 18 from scratch on the full CIFAR-10 data set on a single GPU with a batch size of 32, 64, 128 and 256 as well as perform distributed training on 8 GPUs with batch sizes ranging from 16 to 20,480. We call this complete CIFAR-10 data the \textit{Full} data set. We also use a pretrained model as a point of comparison and find it helps identify features effectively thus saving time and computation.

\subsubsection{Extract features using CNN} 
\begin{figure}[h]
    \centering
    \includegraphics[width=0.4\textwidth]{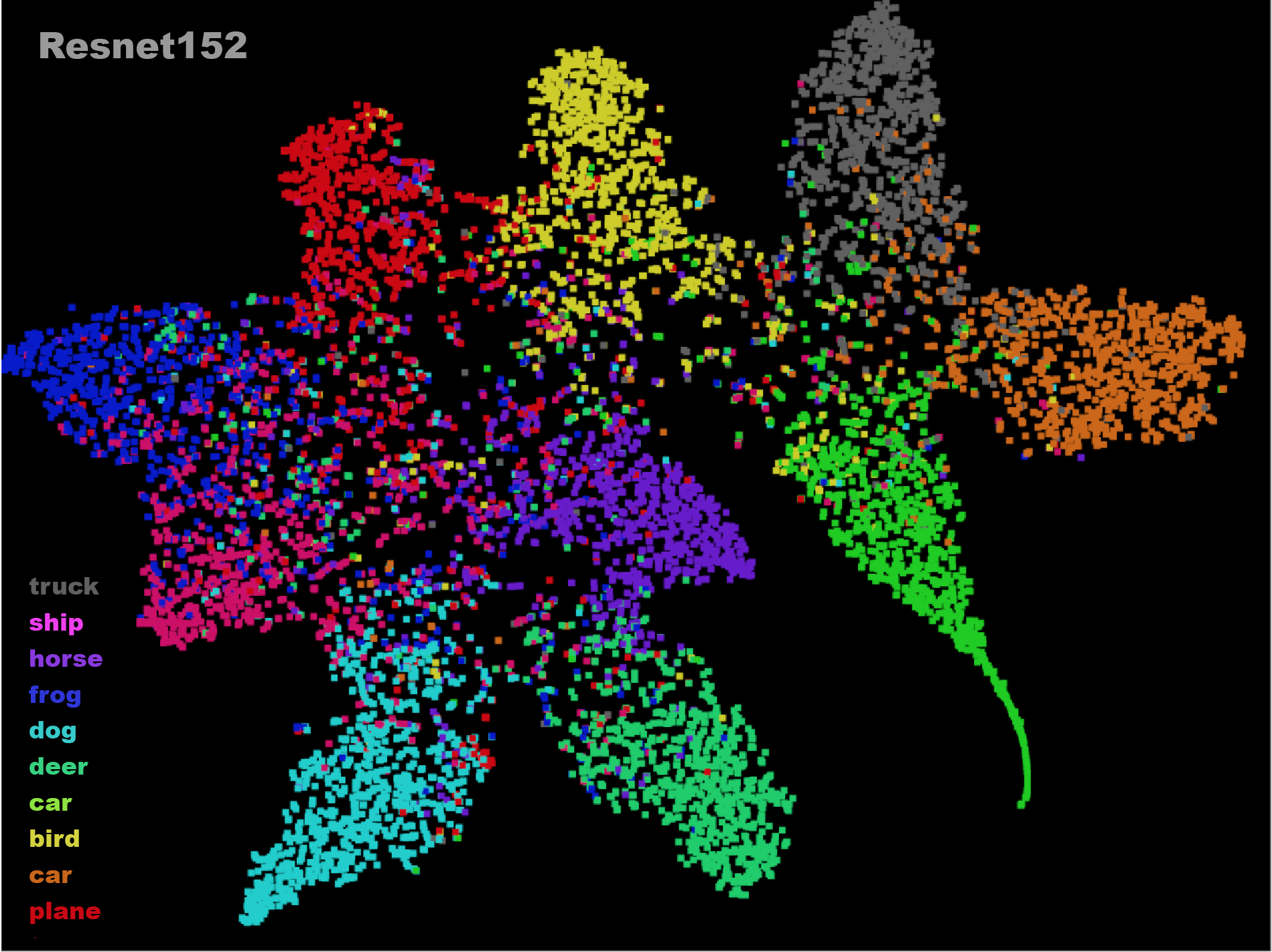}
    \caption{Visualizing output of extracted features from the fine-tuned ResNet-152 using t-SNE.}
    \label{fig:tSNE}
\end{figure}

Next, we focus on the last fully connected layer to extract features using the model as seen in Figure~\ref{fig:tSNE} \cite{dolbdtDNN}. There are 512 neurons in the pen-ultimate layer of ResNet-18 and ResNet-34 models and hence will produce 512 features whereas ResNet-50, 101, and 152 will produce 2048 features.

\subsubsection{Reduce dimensionality using t-SNE} We further reduce the dimensionality of these features to 2 dimensions using t-SNE (and to also help visualize them.) The t-SNE visualization using ResNet-152 of CIFAR-10 test set for all classes shows how the model has learnt to classify the CIFAR-10 data set effectively. We now use the same model to visualize the training data set and apply the DBSCAN clustering technique to classify noise in the next step.

\subsubsection{Clustering to classify and filter data} 
\begin{figure}[ht]
    \centering
    \includegraphics[width=0.4\textwidth]{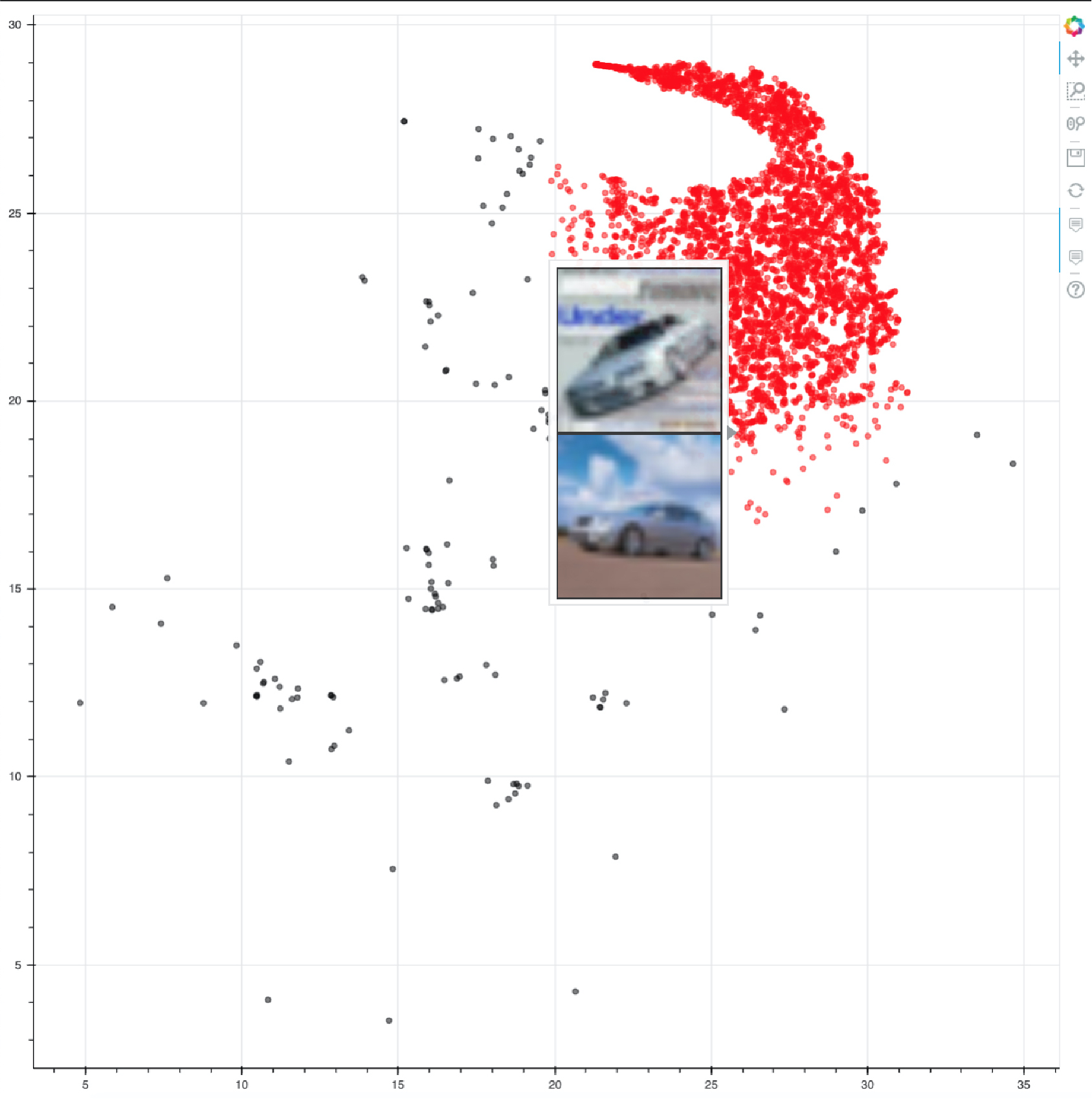}
    \caption{Applying DBSCAN to the t-SNE output, noise shown in black (Automobile).}
    \label{fig:dbscan}
\end{figure}

Next, we cluster the training data set using DBSCAN and filter out the data points (images) classified as noise. (These appear as grey dots in Fig~\ref{fig:dbscan} - source code is available in \cite{dolbdtDNN}. Our code includes an interactive visualizer for t-SNE and DBSCAN which can be used to glance over the images classification - for user oversight.) We remove the images classified as noise from the CIFAR-10 data set and re-purpose the remaining images to train a smaller network. We call our new data set \textit{Network filtered} data.

Different networks (ResNet-18, 34, 50, 101 or 152) achieve different levels of data reduction as shown in Table ~\ref{tb:red}.
\begin{table}[]
\centering
\caption{Amount of data reduction}\label{tb:red}
\begin{tabular}{ |p{4cm}||p{3.8cm}|  }
 \hline
 Network used for feature extraction & Filtered data size (reduction) \\
 \hline\hline
 ResNet-152 & 98.76\% or 49380 images (1.24\%) \\
 \hline
 ResNet-101 & 94.87\% or 47438 images (5.12\%) \\
 \hline
 ResNet-50 & 92.66\% or 46333 images (7.34\%) \\
 \hline
 ResNet-34 & 93.45\% or 46725 images (6.55\%) \\
 \hline
 ResNet-18 & 93.79\% or 46895 images (6.21\%) \\
 \hline
\end{tabular}
%\vspace{0.1cm}
\end{table}   

\subsubsection{Train a smaller network} Finally, we use the Network filtered data from ResNet-101 to train ResNet-18.
Our rationale in using ResNet-101 is to strike a balance between accuracy and data reduction. Using ResNet-101 to reduce data shows promising results and strikes a balance between data reduction and accuracy.

\section{Empirical Observations} \label{sec:results}

\subsection{Batch Size Considerations} 
We train different ResNet-* models on CIFAR-10 data set for 50 epochs using the SGD optimizer and empirically establish for a wide range of batch sizes (from 16 to 20,480) that model accuracy is not linearly correlated with the batch size. Our observations suggest that there is a preferred batch size for this class of models at which the highest accuracy will be achieved at different learning rates.

\begin{figure}[h]
    \centering
    \includegraphics[width=0.45\textwidth]{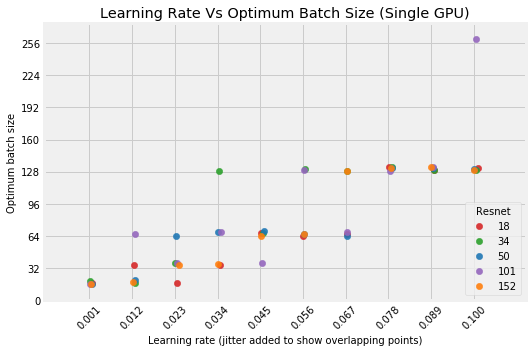}
    \caption{Learning rate vs. Preferred batch size for single GPU}
    \label{fig:optimBS}
\end{figure}

\begin{figure*}[h]
    \centering
    \includegraphics[width=18cm]{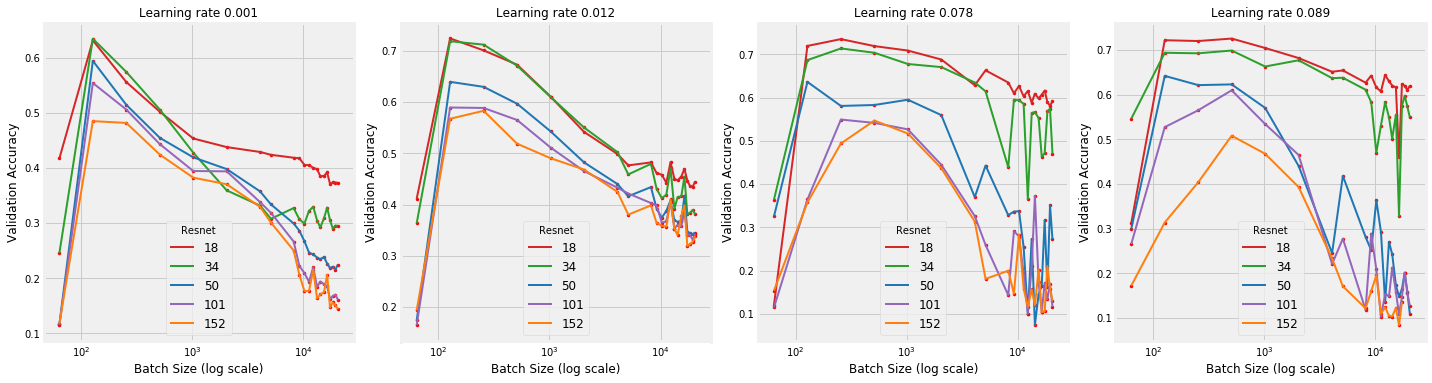}
    \caption{Large batch size vs. Accuracy for different lr. Accuracy increases until it reaches an optimum batch size and then starts to gradually decrease.}
    \label{fig:lbtr}
\end{figure*}

We start with a set of rounds of single GPU training then move to multi GPU distributed training to verify this behavior. This study empirically identifies that there exists an optimal batch size as seen in Figure~\ref{fig:optimBS} at which the highest accuracy is achieved and beyond which accuracy drops gradually, as shown in Figure~\ref{fig:lbtr}, as the batch size increases. This low accuracy for smaller batch sizes applies to batch normalization computation and is explored further in \cite{1804.07612}. Briefly, batch normalization (BN) addresses the problem of {\it{internal covariate shift}} in the neural network by reducing the dependency of the distribution of the input activations of each layer on all the preceding layers. Let us consider batch $X_b$ from $(2)$ of size of $b$

$$\mu_{X_b}=\frac{1}{b}\sum_{i=1}^{b}x_i, \quad \textrm {and} \quad
\sigma_{X_b}^2=\frac{1}{b}\sum_{i=1}^{b}(x_i-\mu_{X_b})^2 \space - \space(7)$$

For a layer of the network with d-dimensional input, \\ $x=(x^{(1)},...,x^{(d)})$, each dimension of its input is then normalized separately,

$${\hat{x}}_{i}^{(k)}={\frac {x_{i}^{(k)}-\hat{\mu}_{X_b}^{(k)}}{\sqrt{\hat{\sigma}_{X_b}^{(k)^2}+\epsilon}}} \space - \space(8)$$

where $k\in[1,d]$ and $i\in[1,m]$; ${\space}$ $\mu_{X_b}^{(k)}$ and $\sigma_{X_b}^{(k)^2}$ are the per-dimension mean and variance, respectively. The normalized values $\hat{x}_i^{(k)}$ are then further scaled and shifted by the learned parameters $\gamma^{(k)}$ and $\beta^{(k)}$

$$y_i^{(k)}=\gamma^{(k)}\hat{x}_i^{(k)}+\beta^{(k)} \space - \space (9)$$

For the case of a convolutional layer, with a feature map of size $p \times q$ and batch size $b$, the sample size for the estimate of $\hat\mu_{X_b}^{(k)}$ and $\hat\sigma_{X_b}^{(k)^2}$ is given by $b\times p \times q$, while for a fully-connected layer the sample size is simply equal to $b$. From (7)  it can be concluded that for very small batch size $b$, the estimation of the batch mean and variance can be very noisy, which may limit the effectiveness of BN in reducing the covariate shift. On the contrary, however, it is well known that too large of a batch size will lead to sub-optimal model convergence and thus poor generalization (although currently it’s not known why this is so). For the convex functions we are trying to optimize, there is an inherent tug-of-war between the benefits of smaller and bigger batch sizes. 

Batch size vs. Accuracy curves for all ResNet models in Figure~\ref{fig:lbtr} tend to run closer and peak together for lower learning rates and diverge as learning rate is increased. This is because for smaller learning rates the optimization step is small for all networks and the resulting variance irrespective of the different loss contours for the networks. Separately, accuracy values for smaller networks like ResNet-18, 34 tend to drop gradually whereas accuracy for bigger networks drops suddenly after achieving the optimal accuracy. This behavior may be attributed to the fact that the larger networks have more complex loss function shapes and hence the optimization step often gets trapped in a local minima.

\subsection{Data Optimized Approach Results} Here we modify the loss landscape by removing noisy data points for ResNet-18 for both small ($<$256 images) as well as large batch sizes (32 images per GPU $\times$ 8 GPUs i.e. 256 images to 1310 images per GPU $\times$ 8 GPUs i.e. 10480 images). We compare our results for the {\it{Network filtered}} data against {\it{Full}} CIFAR-10 data set and {\it{Randomly filtered}} data. We experimented using multiple runs of different weight decay (L2 reg) values of 0.005, 0.01, 0.012 and 0.0135 for small batch sizes and multiple runs of same weight decay value of 0.01 for large batch sizes.

\subsubsection{Small batch training}
\begin{figure*}[h]
    \centering
    \includegraphics[width=0.93\textwidth]{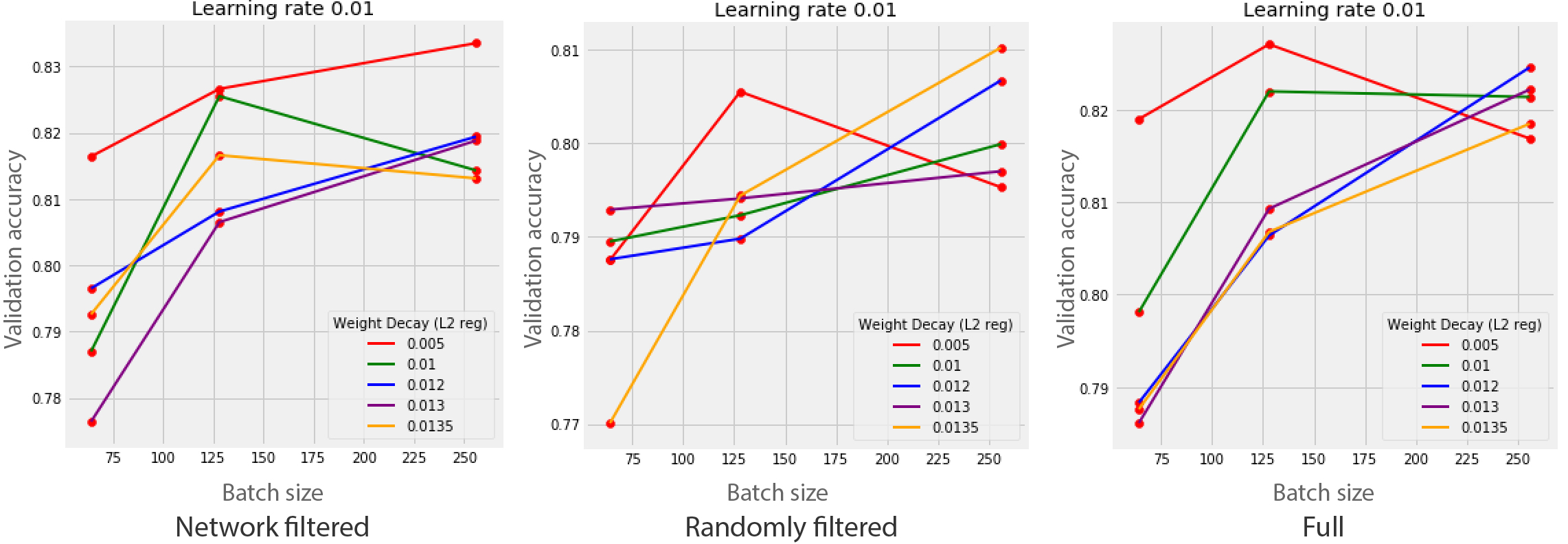}
    \caption{Small batch size vs. Accuracy for network filtered, randomly filtered and full CIFAR-10 data sets.}
    \label{fig:sbtr}
\end{figure*}

For a batch size of 256, as shown in Figure ~\ref{fig:sbtr}, the network achieves top-1 accuracy of 83.2\% at weight decay (L2 reg) of 0.005 on {\it{Network filtered}} data as compared to 82.6\% at weight decay of 0.012 on {\it{Full}} data and 80.9\% at weight decay of 0.0135 on {\it{Randomly filtered}} data. However, as the batch size is reduced, the accuracy of the {\it{Network filtered}} data decreases. For batch size of 64 top-1 accuracy on {\it{Network filtered}} data is 81.8\% at weight decay of 0.005 as compared to 79.2\% for {\it{Randomly filtered}} data at weight decay of 0.012 and for batch size of 128 top-1 accuracy on {\it{Network filtered}} data is 82.7\% at weight decay of 0.005 as compared to 80.5\% for {\it{Randomly filtered}} data at weight decay of 0.005. Accuracy on {\it{Network filtered}} data always remains higher than on {\it{Randomly filtered}} data which is to be expected because the network filtered data is intrinsically biased towards more accurate classification input data elements.

The neural network does better on {\it{Network filtered}} data does than both {\it{Randomly filtered}}  and {\it{Full}} data for smaller regularization values. Thus, we use the lowest regularization value of 0.005 for doing large batch size analysis which we shown in Figure ~\ref{fig:flbtr} and discuss next.

\subsubsection{Distributed large batch training}
\begin{figure}[ht]
    \centering
    \includegraphics[width=0.5\textwidth]{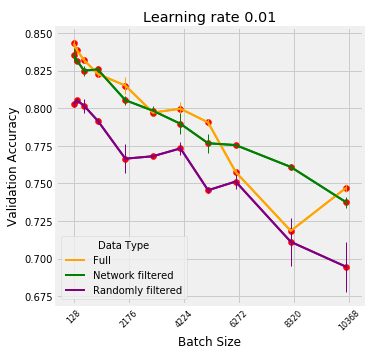}
    \caption{Large batch training accuracy for Full CIFAR-10, Randomly filtered and Network filtered data (Avg. of 4 runs). Network filtered data gives at least the same accuracy as full data and outperforms randomly filtered data.}
    \label{fig:flbtr}
\end{figure}

Our data optimized strategy to improve accuracy by manipulating the loss landscape shows promising results for batch sizes larger than 6k as seen in Figure~\ref{fig:flbtr}. For an average of 4 runs and batch sizes of 6272 to 10368, the neural network consistently achieves a higher top-1 accuracy on {\it{Network filtered}} data than on {\it{Full}} data. For a batch size of 6272, the network achieves top-1 accuracy of 77.5\% on {\it{Network filtered}} data and 75.2\% on {\it{Full}} data.
When compared to training on {\it{Randomly filtered}} data the network consistently achieves a higher accuracy for all batch sizes on {\it{Network filtered}} data.
However, for smaller than 6k batch sizes the accuracy curve on {\it{Network filtered}} data is well within the error bars of the accuracy curve on {\it{Full}} data so we can conclude that the model performs as well on the {\it{Network filtered}} data as it performs on {\it{Full}} data.
Also, as expected the variance in accuracy is higher for larger batch sizes because the larger forward optimization step tends to wiggle around the global minima. A likely cause is that the lack of data points removed is being compensated by the transformations applied to the remaining data points.

%Can we give some hypothesis why  this could be happening based on some assumption???
%Law of large numbers that's why removing a data points doesn't make difference for large batch sizes.
%Data points that are being removed are being compensated by the transformations applied to the current data points.

\subsection{Communication Characteristics and Performance} 
Besides the data-optimized reduction resulting in improved convergence discussed in the above sections, the communication performance involved in large batch training also plays an important role with respect to time-to-solution. The data parallel approach, where the global batch size is divided among model replicas, is one of the leading approaches to scale up. (The other approaches include model parallel and parallel hyper-parameter search.) 

\begin{figure*}[t]
\centering
\includegraphics[width=0.4\textwidth]{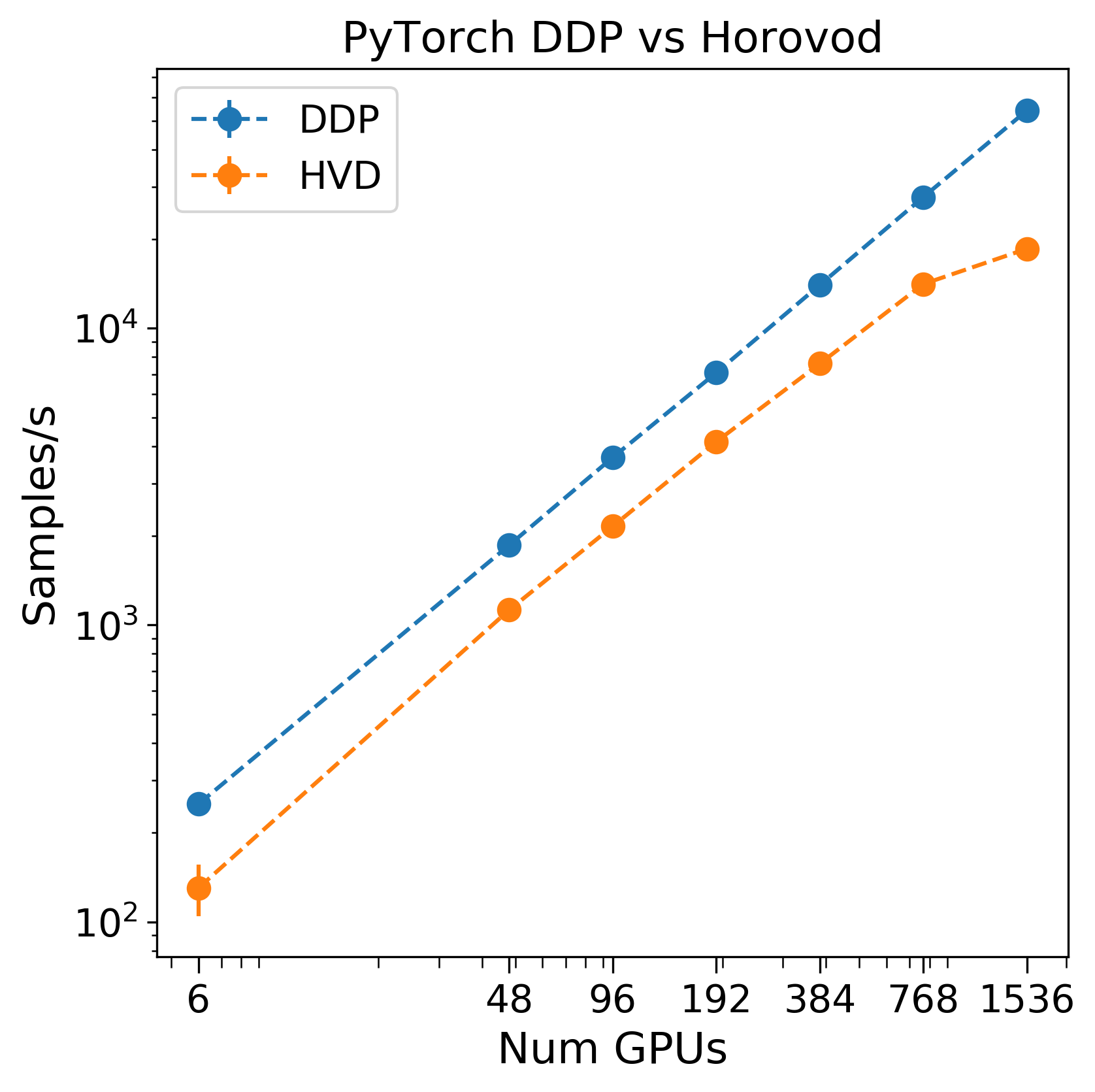}
\includegraphics[width=0.5\textwidth]{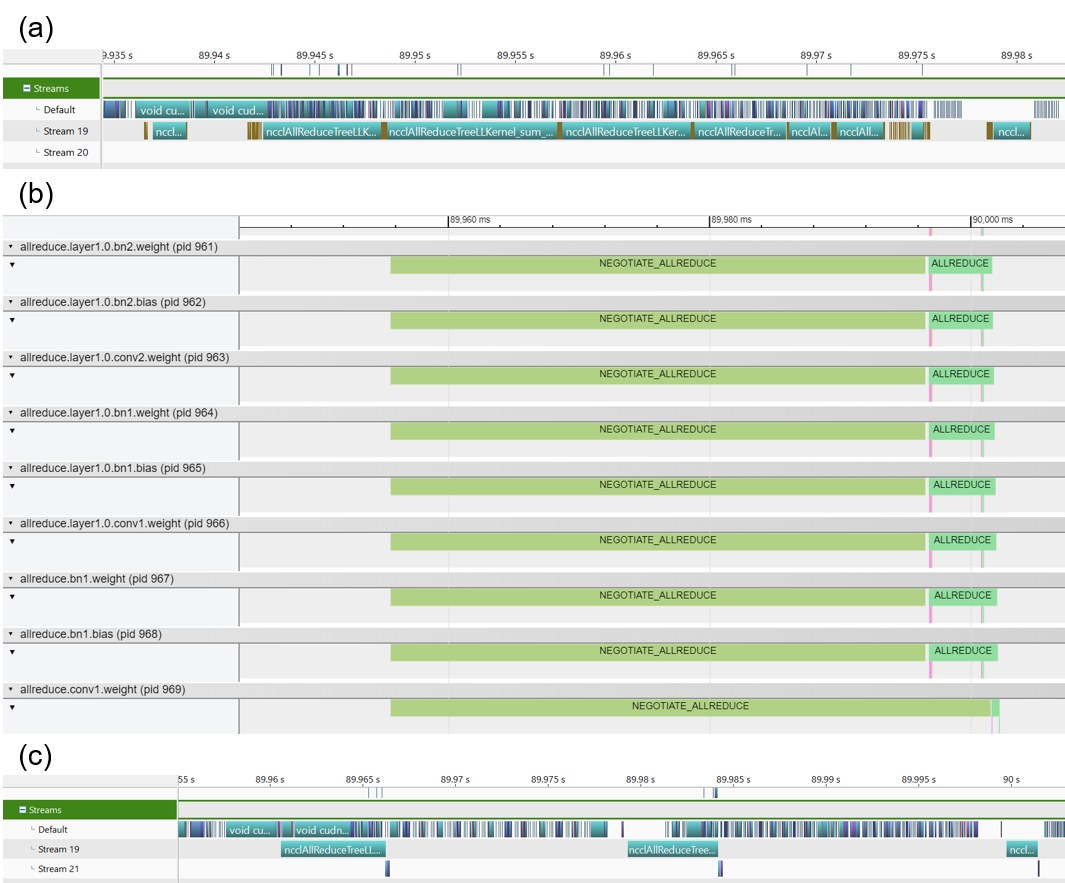}
\caption{(left) Data parallel scaling (effective batch size is linearly proportional to the number of GPUs) of PyTorch with built-in distributed data parallel (DDP) library vs third-party Horovod library, both using NCCL communication backend. (right) (a) nvprof profiling of a typical step in PyTorch with Horovod; (b) Horovod timeline tracing of a step; (c) nvprof profiling of a typical step in PyTorch with DDP. }
\label{fig:dlbt}
\end{figure*}

In Figure~\ref{fig:dlbt}, we compare the 2 most popular distribution libraries for data parallel using PyTorch: 1) Horovod and 2) Distributed Data Parallel (DDP). Horovod, a  third-party plugin, supports a novel communication pattern in which tensors ready to be "allreduced" are placed into a queue (controlled by the buffer size and cycle time parameters) first, and depending on which set of tensors are ready the communications are scheduled dynamically. As shown in Figure~\ref{fig:dlbt}(a) and (b), for one batch update, there is a long negotiation phase to fill up the tensor queue and the communication is divided into several allreduce calls to maximize the overlap between communication and computation (see Figure~\ref{fig:dlbt}(a) 2 kernel streams). However, the overhead of dynamic queuing undermines its benefit, and the scaling deteriorates near 1000 GPUs.  DDP, on the other hand, is a built-in library within PyTorch which handles the tensor reduction in static order and can also hide some communication cost by overlapping communication and computation (see Figure~\ref{fig:dlbt} (c)). For the scaling comparison in Figure~\ref{fig:dlbt}, we use default parameters for buffer size and cycle time for Horovod, default message counts for DDP. Tuning Horovod parameters for 96 GPUs presents only marginal improvements and DDP demonstrates an overall superior scaling for large batch distributed training of ResNet models.              

\section{Conclusion} \label{sec:conclusion}

Large-scale deep neural networks require a large amount of computation to converge to acceptable testing accuracy. Research in efficient and scalable training mostly focuses on efficiently distributing the data to perform data-parallel computation on multiple nodes. 
%Model parallelism and Data parallelism are two approaches which are currently used in practice for distributed deep learning. 
However, data communication between machines in the large-scale system becomes the bottleneck of the system throughput. While using a large mini-batch size can improve the scalability and throughput of the system, it is hard to keep the generalization gap small when data is divided into batches. In this study, we empirically approach this problem from the input data perspective and propose a data optimized strategy for large batch distributed training. We use traditional machine learning techniques (t-SNE and DBSCAN) to refine data quality and remove data points that may be deemed to appear as noise in feature space. We find that for certain large batch sizes of about 6k and above we achieve better performance, and for batch sizes smaller than 6k, we achieve similar accuracy compared to when a model is trained on all the data. For large-batch training in practice, we find that that the PyTorch built-in distributed data parallel messaging library scales better than Horovod for ResNet models trained on up to 1536 V100 GPUs.   

With regards to future extensions of this work, it is worth considering performing hyper-parameter tuning for the DBSCAN and t-SNE algorithms instead of using the default values and using the technique recursively as described in Figure~\ref{fig:process}. This will give us a more robust clustering and filtering on the data set. Different sampling techniques like stratified grid sampling of the sample data points may prove sufficient instead of applying DBSCAN to the entire t-SNE output to remove noisy data. Our work lends itself to conversion into an interactive tool to visualize and investigate the working of a model using such a methodology which can facilitate reproducibility and model transparency. We find that our technique works with a pre-trained ResNet-101 used as a feature extractor as well, instead of training a model from scratch. We intend to expand this work using other pre-trained networks in the future. These techniques are likely to be particularly effective in a variety of domain areas such as  anomaly detection, data compression, and data filtering.


\section*{Acknowledgment}
This research was sponsored by and used resources of the Oak Ridge Leadership Computing Facility (OLCF), which is a DOE Office of Science User Facility and the Compute and Data Environment for Science (CADES) at the Oak Ridge National Laboratory supported by the U.S. Department of Energy under Contract No. DE-AC05-00OR22725.



%\bibliographystyle{IEEEtran}
%\bibliography{bibliography.bib}

\printbibliography %Prints bibliography

\end{document}